\documentclass[letterpaper, 10 pt, conference]{ieeeconf}  

\IEEEoverridecommandlockouts                              
\usepackage[utf8]{inputenc}
\usepackage[T1]{fontenc}
\usepackage{graphicx}
\usepackage{caption}
\usepackage{lipsum}
\usepackage{amsmath,amsfonts,amssymb}
\usepackage{algorithm,algorithmic}
\usepackage{wrapfig}
\usepackage{hyperref}
\usepackage{multirow}
\usepackage{xcolor}

\title{\LARGE \bf
Differentiable Topology Estimating from Curvatures for 3D Shapes
}

\author{ \parbox{3 in}{\centering Yihao Luo\thanks{*This work was not supported by any organization sadsjjjjjjjjjjjjj}\\
        Imperial College London\\
        {\tt\small y.luo23@imperial.ac.uk}}
}

\begin{document}
\twocolumn[{
\renewcommand\twocolumn[1][]{#1}
\maketitle
\begin{center}
    \captionsetup{type=figure}
    \includegraphics[width=\textwidth]{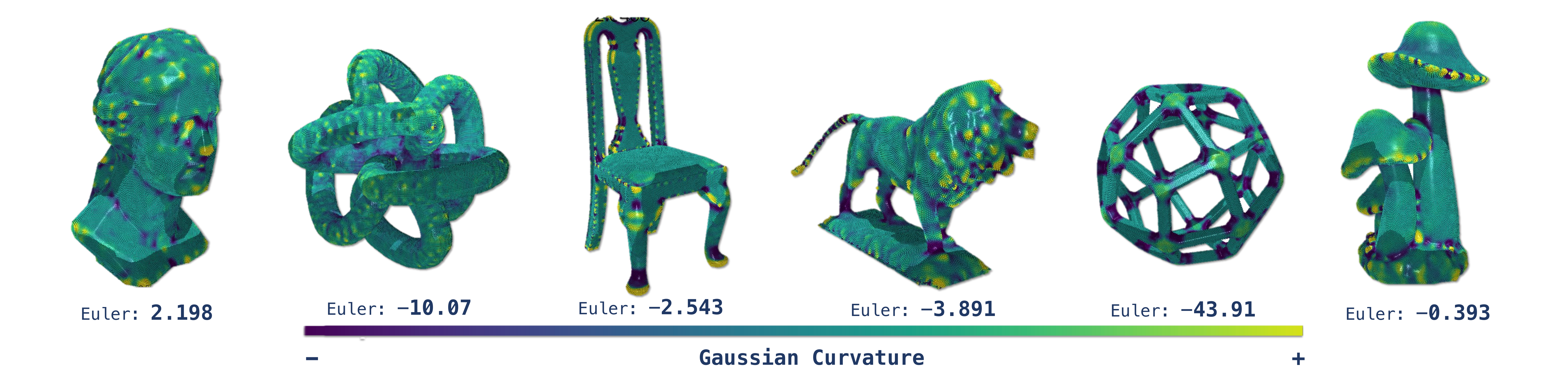}
    \captionof{figure}{\small 
    Our method provides a novel approach to computing the Euler characteristic of 3D shapes in a differentiable manner by analyzing point clouds sampled from any 3D representations. The process begins by calculating Gaussian curvature. followed by a numerically stable integration step that yields an initial estimate of the Euler characteristic. The curvature is visualized as the color map on the 3D models. The ground truth Euler characteristics are $2, -10, -2, -3, -44, 0$ from left to right, and the estimated values are shown under each 3D object. }
    \label{fig:topologyestimation}
\end{center}
}]

\begin{abstract}
In the field of data-driven 3D shape analysis and generation, the estimation of global topological features from localized representations such as point clouds, voxels, and neural implicit fields is a longstanding challenge. This paper introduces a novel, differentiable algorithm tailored to accurately estimate the global topology of 3D shapes, overcoming the limitations of traditional methods rooted in mesh reconstruction and topological data analysis. The proposed method ensures high accuracy, efficiency, and instant computation with GPU compatibility. It begins with an efficient calculation of the self-adjoint Weingarten map for point clouds and its adaptations for other modalities. The curvatures are then extracted, and their integration over tangent differentiable Voronoi elements is utilized to estimate key topological invariants, including the Euler number and Genus. Additionally, an auto-optimization mechanism is implemented to refine the local moving frames and area elements based on the integrity of topological invariants. Experimental results demonstrate the method's superior performance across various datasets. The robustness and differentiability of the algorithm ensure its seamless integration into deep learning frameworks, offering vast potential for downstream tasks in 3D shape analysis.
\end{abstract}    
\section{Introduction}

In recent years, 3D shape analysis and generation have become increasingly popular research topics in computer vision, computer graphics, and machine learning. Many impressive studies and applications have been developed to achieve 3D object recognition, reconstruction, and generation. However, the data-driven 3D shape analysis and generation still face many challenges regardless of the kind of localized data representation used, from point clouds, occupancy or signature distance fields (SDF) \cite{genova2019learning,park2019deepsdf,marschner2023constructive,zhang2021learning}, to neural radiance fields (NeRF) \cite{mildenhall2021nerf,muller2022instant,poole2022dreamfusion} and Gaussian Splattings (GS) \cite{kerbl20233d,wuc2024recent,chen2024survey,huang20242d}. One essential challenge is that existing methods are usually locally focused and lack consistency and coherence in terms of global view, which causes several deformities and artifacts in the generated 3D shapes. Although researchers have made great efforts to improve the global coherence of 3D shapes via more advanced architectures and supervision mechanics in deep learning \cite{park2019deepsdf,genova2019learning,kerbl20233d,mildenhall2021nerf}, few studies have focused on the global topology of 3D shapes and its estimation from the localized data representations in a differentiable manner. In this paper, we propose a differentiable algorithm to estimate the global topology of 3D shapes, which can be easily integrated into deep learning frameworks.

The global topology of 3D shapes is a fundamental property that characterizes the shape's connectivity and the number of holes. The main reason is that the 3D shape contains rich and essential information from its global topology, which is usually hard to learn by a locally focused optimization from deep learning. Therefore, a differentiable algorithm that can accurately estimate the global topology of 3D shapes with modalities adaptivity is highly demanded. On the other hand, efficient topological estimation is also challenging in traditional computer graphics and computational geometry, let alone the requirement for differentiability. Established methods to analyze the topology of discrete 3D shapes include the Reeb graph \cite{pascucci2007robust}, Morse theory \cite{milnor1963morse}, and persistent homology 
\cite{edelsbrunner2008computational, edelsbrunner2002topological, carlsson2009topology} from topological data analysis (TDA). However, current methods have inherent limitations in differentiability and computational efficiency, making them difficult to integrate into the deep learning frameworks. 

To achieve globality and differentiability simultaneously, we propose a novel algorithm to estimate topology based on the estimation of the local curvatures and its stable integration, inspired by the great idea of Gauss-Bonnet theorem \cite{wu2008historical}. The Gauss-Bonnet theorem is a fundamental result in differential geometry that establishes a relationship between the local curvature and the global topology of a surface. It states that the integral of the Gaussian curvature $K$ over a surface (manifold) ${M}$ is equal to $2\pi$ times the Euler characteristic $\chi_{{M}}$, i.e.,
\begin{equation}\label{eq:gaussbonnet}
\frac{1}{2\pi}\int_{{M}} K dA = \chi_{{M}},
\end{equation}
where $dA$ is the integral area element.

From now on, the question turns to how to estimate the local curvatures and integrate them in a differentiable and stable way.  Efficient estimation of (Gaussian/mean/principal) curvature is also an essential but difficult problem. Taking the point cloud as an example (unordered point sets contain the least information and can be easily extracted from other modalities), several methods have been proposed to estimate the curvature. \cite{merigot2010voronoi} introduced the Voronoi covariance measure (VCM) to estimate the curvature for noised point clouds. Robust methods varied from local statistics on the sampled neighborhoods are the prevalent alternatives to estimate directional curvatures while coping with noise and irregular sampling, such as \cite{wardetzky2007discrete} involving least squares and \cite{panozzo2010efficient} adopting local re-weighting. The classic paper \cite{taubin1995estimating} defined the discrete Weingarten map in an integration formula on meshed and \cite{cao2021efficient} adopted it to estimate the principal curvature on point clouds. \cite{berkmann1994computation} proposed a method for computing the covariance matrices of normal vectors to approximate the shape operator (equivalent to the Weingarten map to some extent). More recently, \cite{kalogerakis2007robust} used robust statistical methods to estimate the curvature tensor of a point cloud by considering normal variation in a neighborhood of a given point. \cite{yang2007direct} derived analytic expressions for computing principal curvatures based on the implicit definition of the moving least squares surface by \cite{amenta2004domain}. \cite{cao2021efficient} formulated the statistical algorithms and gave out the convergence analysis. However, none of them shows sufficient robustness and accuracy to support the integration for global topology where error accumulation occurs. 

We tend to estimate the curvature by the Weingarten map following the idea of \cite{cao2021efficient}, but especially emphasize the self-adjoint property of the Weingarten map, which is crucial for the stability of the curvature estimation. Moreover, our method will be specifically designed to avoid non-differentiable operations and non-stable computations, which would probably cause the gradient explosion or vanishing during the back-propagation in deep learning. Subsequently, the curvatures will be integrated over tangent differentiable Voronoi elements \cite{aurenhammer1991voronoi}, where we obtain the initial estimation of the Euler characteristic and genus. Finally, a self-optimization mechanic leveraging the integrity of topological invariants is implemented to rectify the local moving frames and area elements, from which the fine estimation of the topology can be achieved. The self-optimization addresses the unreliability of the initial normal estimation and the area elements from local statistics. The occupancy information of spacial points contained in occupancy or SDF will be coded into the winding number and play the role of regularization in the differentiable form inspired by \cite{xu2023globally}.




\section{Methods}

In this section, we present our methodology, starting with the estimation of moving frames and local area elements. We then analyze the self-adjoint Weingarten map and propose a highly robust approach for estimating Gaussian curvature. Finally, we incorporate curvature estimates with a self-optimization technique that refines local frames and area elements, optimizing them in alignment with the integrity-well loss to preserve the global topology.

\subsection{Local Tangent Voronoi Area}
Consider a point cloud \(\mathcal{P} := \{\mathbf{p}_i\}_{i = 1}^N\) lying on an underlying surface \({M} \subseteq \mathbb{R}^3\). For each point \(\mathbf{p}_i\), a moving orthonormal frame \(\{\mathbf{t}_i, \mathbf{t}^{\prime}_i, \mathbf{n}_i\}\) is defined by a group of basis spanning the $3$-dimensional linear space originating from \(\mathbf{p}_i\), where \(\mathbf{t}_i\) and \(\mathbf{t}^{\prime}_i\) span the tangent space $T_{\mathbf{p}_i}M$, and \(\mathbf{n}_i\) is the normal vector. 

In a discrete setting, the orthonormal frame at a point \(\mathbf{p}_i\) can be estimated via local principal component analysis (PCA) of its \(k\)-nearest neighbors (kNN). Let \(\mathcal{N}_i = \{\mathbf{p}_{ij}\}_{j=1}^{k}\) denote the \(k\)-nearest neighbors of \(\mathbf{p}_i\), organized as a \(k \times 3\) matrix. 
Then eigenvectors of \(\operatorname{cov}(\mathcal{N}_i)\) corresponding to the first two largest eigenvalues represent the tangent vectors \(\mathbf{t}_i\) and \(\mathbf{t}^{\prime}_i\), while the eigenvector associated with the smallest eigenvalue defines the normal vector \(\mathbf{n}_i\). 

Since the exact eigenvalues are not needed for subsequent computations, we perform local PCA using a normalized version of the covariance matrix, \(\operatorname{cov}(\mathcal{N}_i)/\mathrm{Tr}(\operatorname{cov}(\mathcal{N}_i))\), where division by the trace \(\mathrm{Tr}(\operatorname{cov}(\mathcal{N}_i))\) helps mitigate scale effects and improves computational stability.

In theory, the local positional offsets \(\mathrm{d}\mathbf{p}_{ij} = \mathbf{p}_{ij} - \mathbf{p}_i\) lie in the tangent space \(T_{\mathbf{p}_i}M\). However, due to sparsity in local neighborhoods, they may not be perfectly tangent to the surface. Practically, \(\mathrm{d}\mathbf{p}_{ij}\) is projected onto the tangent space using inner products with \(\mathbf{t}_i\) and \(\mathbf{t}^{\prime}_i\). This projection yields a 2-dimensional local tangent coordinate system for \(\mathcal{N}_i\), i.e.,
\begin{equation}\label{local tangent coordinate system}
\mathrm{d}\widetilde{\mathbf{p}}_{ij} = \mathrm{d}\mathbf{p}_{ij} [\mathbf{t}_i, \mathbf{t}^{\prime}_i],
\end{equation}
where \(\mathrm{d}\widetilde{\mathbf{p}}_{ij}\) forms a \(k \times 2\) matrix representing a 2D local point set on the tangent plane.

\begin{wrapfigure}{r}{0.2\textwidth}
\includegraphics[width=0.2\textwidth]{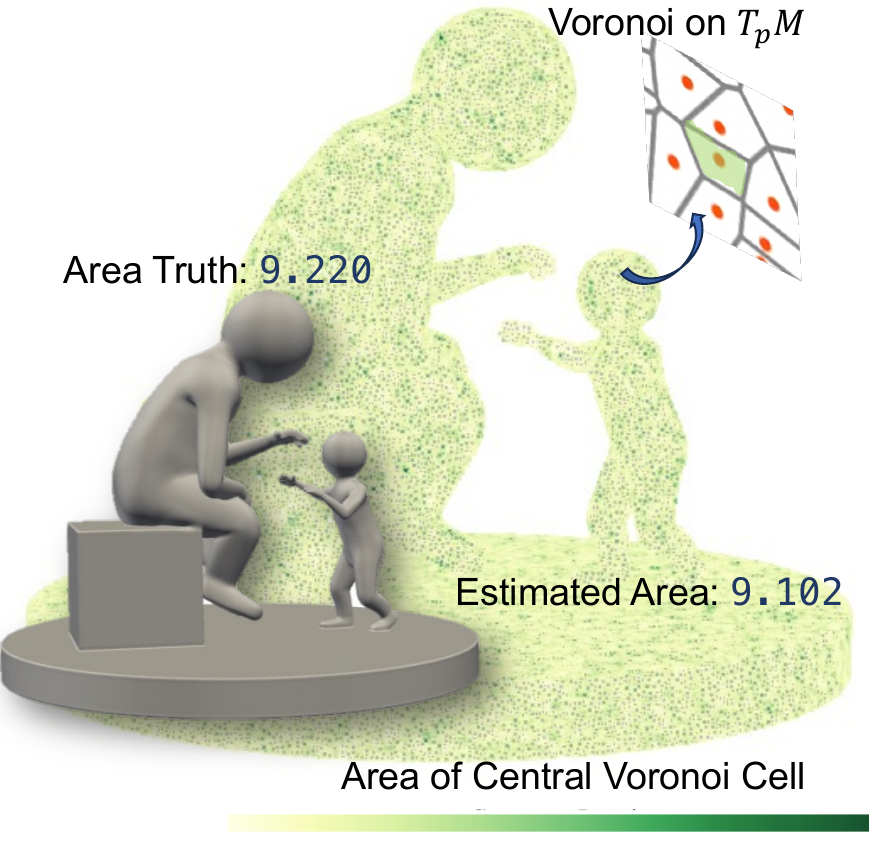}
\caption{\small Tangent Voronoi Diagram for estimating areas.}
\label{fig:Voronoi}
\end{wrapfigure}

The local area element at point \(\mathbf{p}_i\) is estimated using a tangent Voronoi diagram constructed on the 2D point set \(\mathrm{d}\widetilde{\mathbf{p}}_{ij}\), as illustrated in Fig. \ref{fig:Voronoi}. The Voronoi diagram \cite{aurenhammer1991voronoi} partitions the entire region into polygonal cells, each associated with a seed point. The central Voronoi cell of \(\mathbf{p}_i\) defines an exclusive region whose area represents the local area element of \(\mathbf{p}_i\).

To enable differentiable computation, the local area element can be estimated using Monte Carlo simulation. For point \(\mathbf{p}_i\), a 2D mesh grid is constructed over a suitably scaled tangent bounding square (scaled by \( \times 1.1\) in practice) encompassing \(\mathrm{d}\widetilde{\mathbf{p}}_{ij}\). This grid is denoted by \(\mathcal{G}_i\). Let \(\mathcal{G}_i^{c}\) denote the grid points within the central Voronoi cell, identified by 1-nearest neighbor (1-NN) searching as follows:
\[
\mathcal{G}_i^{c} = \left\{ v \in \mathcal{G}_i \mid d(v, \mathbf{p}_i) \leq d(v, \mathcal{N}_i) \right\},
\]
where \(d(v, \mathbf{p}_i)\) is the distance from grid point \(v\) to \(\mathbf{p}_i\), and \(d(v, \mathcal{N}_i)\) is the minimum distance to any neighbor in \(\mathcal{N}_i\). The area element \(A_i\) is approximated by the Monte Carlo ratio:
\begin{equation}\label{area element}
    A_i \approx \frac{\# \mathcal{G}_i^{c}}{\# \mathcal{G}_i} A_{bbx},
\end{equation}
where \(\#\) denotes the number of points in a finite set, \(A_{bbx}\) is the area of the bounding square, and the ratio represents the proportion of grid points in the central Voronoi cell. The summation (discrete integration of the constant function $f \equiv 1$) over all local Voronoi area elements provides a differentiable estimate of the total area of the 3D shape.

Fig. \ref{fig:Voronoi} demonstrates tangent Voronoi area estimation on a 10K point cloud of the model "Father's Strength"\footnote{The 3D model "Father's Strength" is available on \href{https://www.turbosquid.com/3d-models/father-s-strength-880248}{TurboSquid}, created by Lurisay, and will be consistently used as the showcase in Section 2.}. The estimated area \(A = 9.1376\) closely approximates the ground truth \(\hat{A} = 9.220\), yielding a relative error of \(0.93\%\). Refer to Section \ref{sec:experiments} for additional experimental results.

\subsection{\textbf{Self-Adjoint Weingarten Fields and Curvatures}}


\begin{figure}[hpt]
\centering
\includegraphics[width=0.35\textwidth]{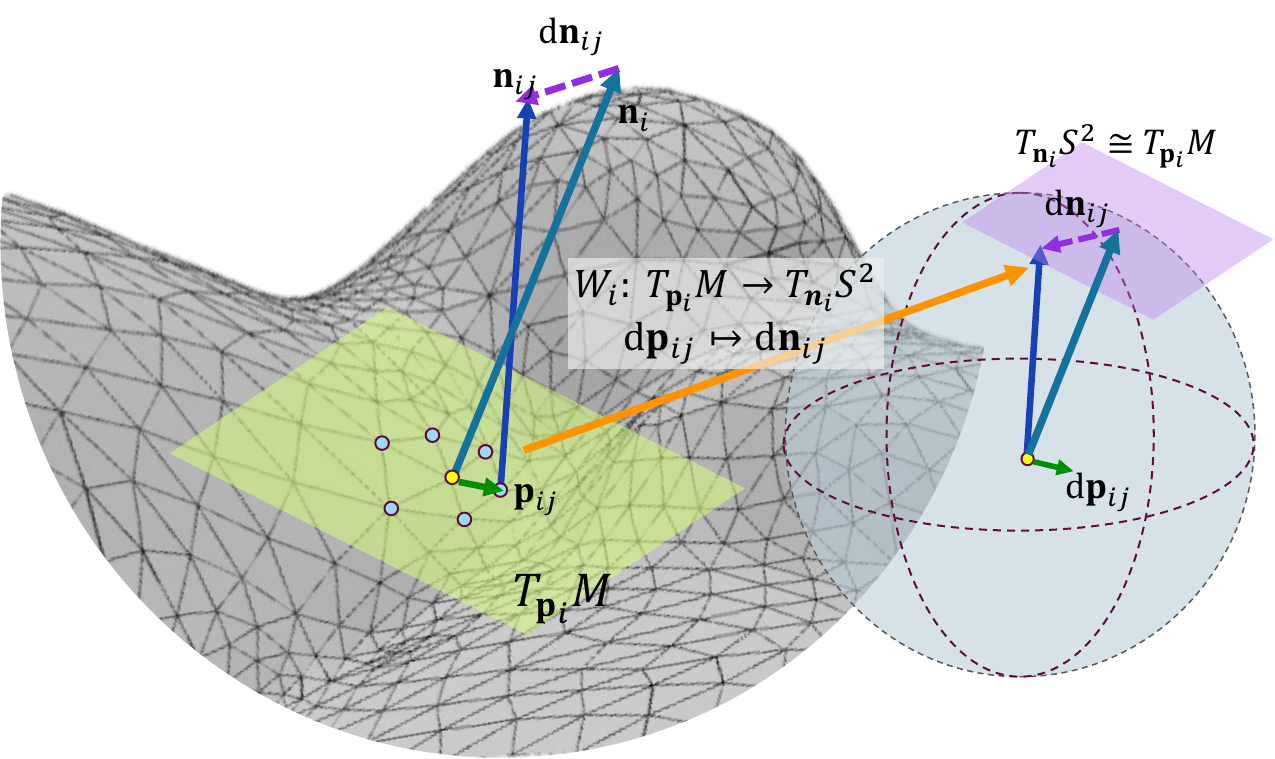}
\caption{ \small Illustration of the Weingarten map for the manifold $M$.}
\label{fig:weingarten}
\end{figure}

With a method to estimate the local area element, we can now proceed to estimate the curvatures of the 3D shape. The Weingarten map is the foundation for estimating curvatures, where two principal curvatures are two eigenvalues of the Weingarten map. The Weingarten map is defined as the derivative of the normal (Gaussian) map. Under the fact that $$\langle \mathbf{n}_i, \mathrm{d}\mathbf{n}_i \rangle = \frac{1}{2}\mathrm{d}\langle \mathbf{n}_i, \mathbf{n}_i \rangle = 0,$$ $\mathrm{d}\mathbf{n}_i$ is tengent to $M$ and the tangent space of $M$ at $\mathbf{p}_i$ is naturally identifed to the tangent space of standard sphere $S^2$ at $\mathbf{n}_i$ , i.e., 
$T_{\mathbf{p}_i}M \cong T_{\mathbf{n}_i}S^2.$
Hence the Weingarten map $W_i$ can be viewed as an endomorphism on $T_{\mathbf{p}_i}M$. In our settings, we have  
\begin{equation}\label{definition of Weingarten map}
  \begin{aligned}
  W_i:\ & T_{\mathbf{p}_i}M \rightarrow T_{\mathbf{n}_i}S^2 \cong T_{\mathbf{p}_i}M,\\
  & W_i(\mathbf{v}) = -\nabla_{\mathbf{v}}\mathbf{n}_i,\ \forall \mathbf{v} \in T_{\mathbf{p}_i}M. \\
  \end{aligned}
\end{equation}
Notice that $W_i: \mathrm{d}\mathbf{p}_{ij} \mapsto \mathrm{d}\mathbf{n}_{ij}$ only holds inside the tangent plane in idealized cases, we approximate the matrix equation $W_i\mathrm{d}\mathbf{p}_{ij}^\intercal = \mathrm{d}\mathbf{n}_{ij}^\intercal$ by their tangent projections  $W_i\mathrm{d}\widetilde{\mathbf{p}}_{ij}^\intercal = \mathrm{d}\widetilde{\mathbf{n}}_{ij}^\intercal$, where \(W_i\) is written as a \(2 \times 2\) matrix. Figure \ref{fig:weingarten} demonstrates the above conceptions. The estimation of $W_i$ attributes to solving the least squares problem:
\begin{equation}\label{eq:leastsquare}
W_i = \arg\min_{W_i}  \|W_i \mathrm{d}\widetilde{\mathbf{p}}_{ij}^\intercal - \mathrm{d}\widetilde{\mathbf{n}}_{ij}^\intercal\|_F^2.
\end{equation}
Traditionally, the solution to Eq. \eqref{eq:leastsquare} is obtained through the pseudo-inverse,
\begin{equation}\label{eq:pseudo-inverse}
W_i = (\mathrm{d}\widetilde{\mathbf{p}}_{ij}^\intercal \mathrm{d}\widetilde{\mathbf{p}}_{ij})^{-1} \mathrm{d}\widetilde{\mathbf{p}}_{ij}^\intercal \mathrm{d}\widetilde{\mathbf{n}}_{ij},
\end{equation}
as implemented in \cite{cao2021efficient}. However, an important property of the Weingarten map is that it is self-adjoint (or Hermitian \cite{chen1999lectures}), i.e., \(W_i = W_i^\intercal\). This property follows from the definition of the Weingarten map \eqref{definition of Weingarten map} and its commutativity with the inner product:
\begin{equation*}
\label{eq:selfadjoint}
\begin{aligned}
  \mathbf{u}^\intercal W_i \mathbf{v} =& -\langle\nabla_{\mathbf{u}}\mathbf{n}_i, \mathbf{v} \rangle = \langle \mathbf{n}_i, \nabla_{\mathbf{u}}\mathbf{v} \rangle \\
= &-\langle \mathbf{n}_i, \nabla_{\mathbf{v}}\mathbf{u} \rangle = -\langle \mathbf{u}, \nabla_{\mathbf{v}}\mathbf{n}_i \rangle \\  
= &\ \mathbf{v}^\intercal W_i \mathbf{u},\quad \forall \mathbf{u}, \mathbf{v} \in T_{\mathbf{p}_i}M,
\end{aligned}
\end{equation*}
where the second and fourth equalities hold due to the orthogonality of the normal vector \(\mathbf{n}_i\) with the tangent vectors \(\mathbf{u}\) and \(\mathbf{v}\). The third equality follows from the Lie bracket 
\[
[\mathbf{v}, \mathbf{u} ] = \nabla_{\mathbf{u}} \mathbf{v} - \nabla_{\mathbf{v}} \mathbf{u},
\]
which lies in the tangent plane and is orthogonal to the normal vector. The self-adjoint property of the Weingarten map guarantees real eigenvalues, which correspond to the principal curvatures. Therefore, obtaining a symmetric solution for Eq. \eqref{eq:leastsquare} is essential for stable curvature estimation.

Combining the linearity of \(W_i\), \(W_i = W_i^\intercal\), and Eq. \eqref{definition of Weingarten map}, we can derive 
\begin{equation}
\label{eq:robustestimationofW}
\begin{aligned}
& {\rm d}\widetilde{\mathbf{p}}_{ij}^\intercal {\rm d}\widetilde{\mathbf{n}}_{ij} + {\rm d}\widetilde{\mathbf{n}}_{ij}^\intercal {\rm d}\widetilde{\mathbf{p}}_{ij} =  \left( {\rm d}\widetilde{\mathbf{p}}_{ij}^\intercal {\rm d}\widetilde{\mathbf{p}}_{ij}\right) W_i + W_i \left( {\rm d}\widetilde{\mathbf{p}}_{ij}^\intercal {\rm d}\widetilde{\mathbf{p}}_{ij}\right),
\end{aligned}
\end{equation}
which defines a Sylvester equation \cite{ward1991general} for the self-adjoint Weingarten map \(W_i\). The algebraic properties of the Sylvester equation are thoroughly described in \cite{luo2021geometric, zhang2019new}, and the solution to Eq. \eqref{eq:robustestimationofW} can be computed using the following Algorithm \ref{alg:robustestimationofW}.


\begin{algorithm}[h!]
  \caption{Solution of the Self-adjoint Weingarten Map}
  \hspace*{0.02in} {\bf Input:} Matrices \( A = {\rm d}\widetilde{\mathbf{p}}_{ij}^\intercal{\rm d}\widetilde{\mathbf{p}}_{ij} \) and 
  \[
    X = {\rm d}\widetilde{\mathbf{p}}_{ij}^\intercal{\rm d}\widetilde{\mathbf{n}}_{ij} +  {\rm d}\widetilde{\mathbf{n}}_{ij}^\intercal{\rm d}\widetilde{\mathbf{p}}_{ij} 
  \]
  \hspace*{0.02in} {\bf Output:} Solution of \( W \) such that \( WA + AW = X \) \\
  \vspace{-0.4cm}
  \begin{algorithmic}[1]
    \STATE \textbf{Eigen-decompose} \( A \): Compute \( A = Q \Lambda Q^T \), where \( Q \in O(n) \) (orthogonal matrix) and \( \Lambda = {\rm diag}(\lambda_1, \dots, \lambda_n) \) contains the eigenvalues \( \lambda_i \).
    \STATE \textbf{Transform} \( X \): Calculate \( C_X := [c_{ij}] = Q^T X Q \).
    \STATE \textbf{Compute} \( E_X \): Define \( E_X = [e_{ij}] \) by setting \( e_{ij} = \frac{c_{ij}}{\lambda_i + \lambda_j} \), provided that \( \lambda_i + \lambda_j \neq 0 \).
    \STATE \textbf{Reconstruct} \( W \): Obtain \( W = Q E Q^T \).
  \end{algorithmic}
  \label{alg:robustestimationofW}
\end{algorithm}

\begin{figure}[hpt]
\centering
\includegraphics[width=0.35\textwidth]{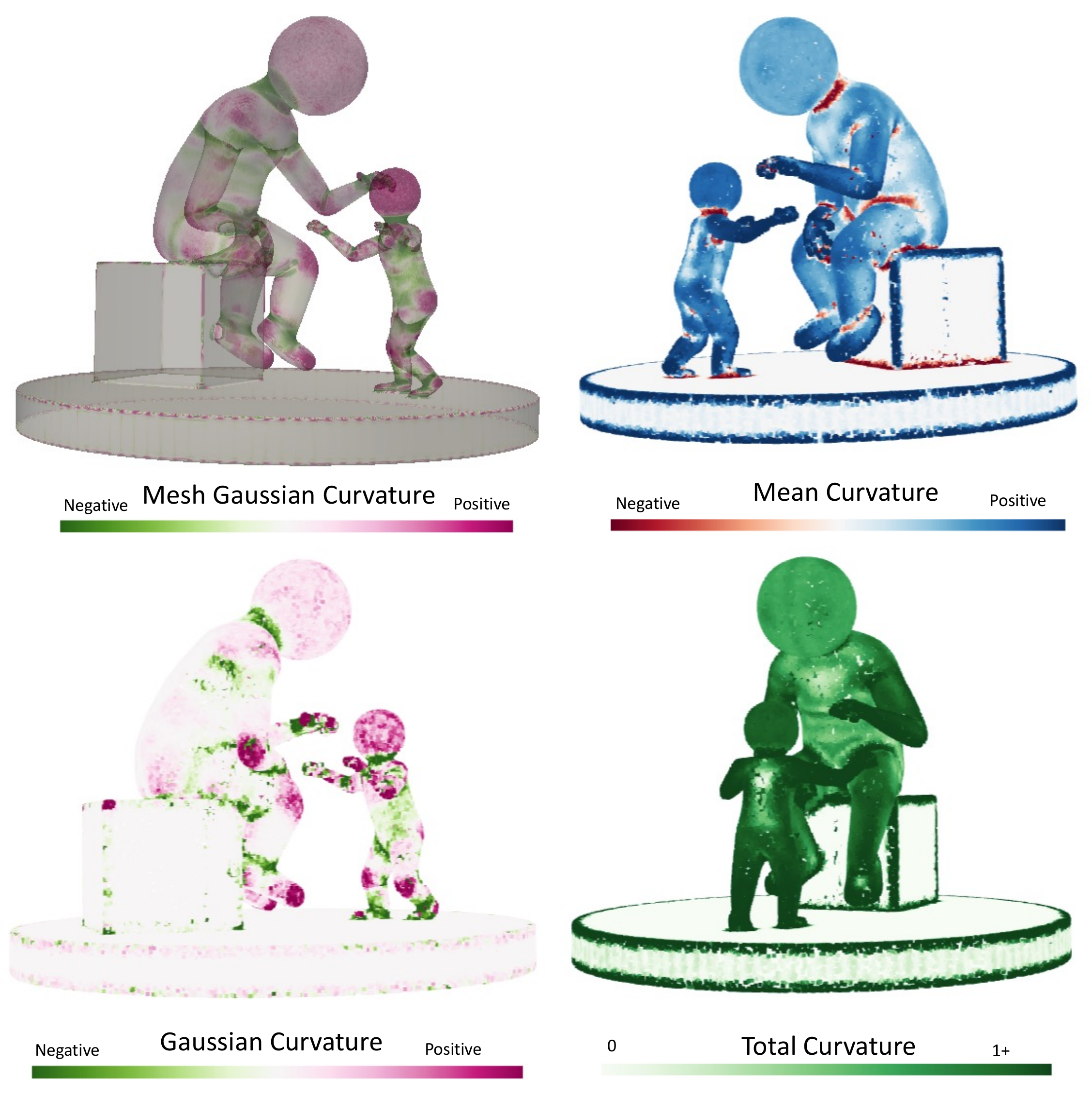}
\caption{ \small The estimation of the Weingarten map and the curvatures.}
\label{fig:curvatureestimation}
\end{figure}

By the eigen-decomposition of the symmetric matrix \(W_i\), the Gaussian curvature is calculated as the product of the principal curvatures equal to the determinate of \(W_i\), 
\begin{equation}\label{Gaussian curvature}
    K_i = \lambda_{\min}(W_i) \lambda_{\max}(W_i) = \det(W_i).
\end{equation}
The mean curvature is given by the average of the principal curvatures:
\[
H_i = \frac{1}{2}(\lambda_{\min}(W_i) + \lambda_{\max}(W_i)) = \frac{1}{2} \operatorname{Tr}(W_i).
\]
Additionally, the total curvature can be derived from the Frobenius norm of the Weingarten map:
\[
F_i = \|W_i\|_F = \operatorname{Tr}(W_i^\intercal W_i).
\]
Figure \ref{fig:curvatureestimation} illustrates the curvature estimation results.

Algorithm \ref{alg:robustestimationofW} exclusively uses differentiable operations, enabling gradient backpropagation. However, multiple steps of eigen-decompositions can be computationally intensive and may lead to gradient explosion. To enhance stability and efficiency, we propose two strategies to simplify Algorithm \ref{alg:robustestimationofW}, particularly for Gaussian curvature estimation.

The first approach assumes that \(W_i\) and \({\rm d}\widetilde{\mathbf{p}}_{ij}^\intercal {\rm d}\widetilde{\mathbf{p}}_{ij}\) commute, implying
\[
W_i {\rm d}\widetilde{\mathbf{p}}_{ij}^\intercal {\rm d}\widetilde{\mathbf{p}}_{ij} = {\rm d}\widetilde{\mathbf{p}}_{ij}^\intercal {\rm d}\widetilde{\mathbf{p}}_{ij} W_i.
\]
This assumption simplifies the Sylvester equation \eqref{eq:robustestimationofW} as follows:
\begin{equation*}
W_i {\rm d}\widetilde{\mathbf{p}}_{ij}^\intercal {\rm d}\widetilde{\mathbf{p}}_{ij} = \frac{1}{2}\left( {\rm d}\widetilde{\mathbf{p}}_{ij}^\intercal {\rm d}\widetilde{\mathbf{n}}_{ij} +  {\rm d}\widetilde{\mathbf{n}}_{ij}^\intercal {\rm d}\widetilde{\mathbf{p}}_{ij} \right).
\end{equation*}
Taking the determinant on both sides yields that
\begin{equation}\label{approximate gaussian curvature}
K_i = \det(W_i) \approx \frac{\det\left( {\rm d}\widetilde{\mathbf{p}}_{ij}^\intercal {\rm d}\widetilde{\mathbf{n}}_{ij} + {\rm d}\widetilde{\mathbf{n}}_{ij}^\intercal {\rm d}\widetilde{\mathbf{p}}_{ij}\right)}{4 \det\left({\rm d}\widetilde{\mathbf{p}}_{ij}^\intercal {\rm d}\widetilde{\mathbf{p}}_{ij} \right)},
\end{equation}
where eigen-decomposition is avoided. The non-negativity and symmetry of the covariance matrix in the denominator help ensure stability, except for singular local distributions, which can be addressed with small perturbations. The second approach involves directly symmetrizing the non-Hermitian solution of Eq.\eqref{eq:leastsquare} by updating 
\begin{equation}\label{eq:symmetrizing}
    W_i \leftarrow \frac{1}{2}(W_i + W_i^\intercal),
\end{equation}
where the $W_i$ on the right-hand side is given by Eq.\eqref{eq:pseudo-inverse}.

\subsection{\textbf{Differentiable Topology Estimation}}

By integrating the differentiable estimation of local area elements and curvatures, we can approximate the global topology of the point cloud. The Euler characteristic \(\chi_M\) is estimated using the discrete Gauss-Bonnet theorem:
\begin{equation}\label{eq:euler}
\chi_M \approx \frac{1}{2\pi}\sum_{i=1}^{N} K_i A_i,
\end{equation}
where \(K_i\) is the Gaussian curvature and \(A_i\) is the area element of point \(p_i\). From the Riemann-Hurwitz formula \cite{chen1999lectures}, the genus \(g(M)\) of the closed surface can be derived as
\begin{equation}\label{Riemann-Hurwitz}
\chi_M = 2 - 2g(M).
\end{equation}

Though, in theory, topological invariants are integers, our estimates yield real numbers due to differentiable calculations and estimation errors. However, this minor discrepancy can works as a regularization mechanism for lower-order geometric features like local offsets, normals, area elements, and tangent frames.

\begin{wrapfigure}{r}{0.0\textwidth}
\includegraphics[width=0.23\textwidth]{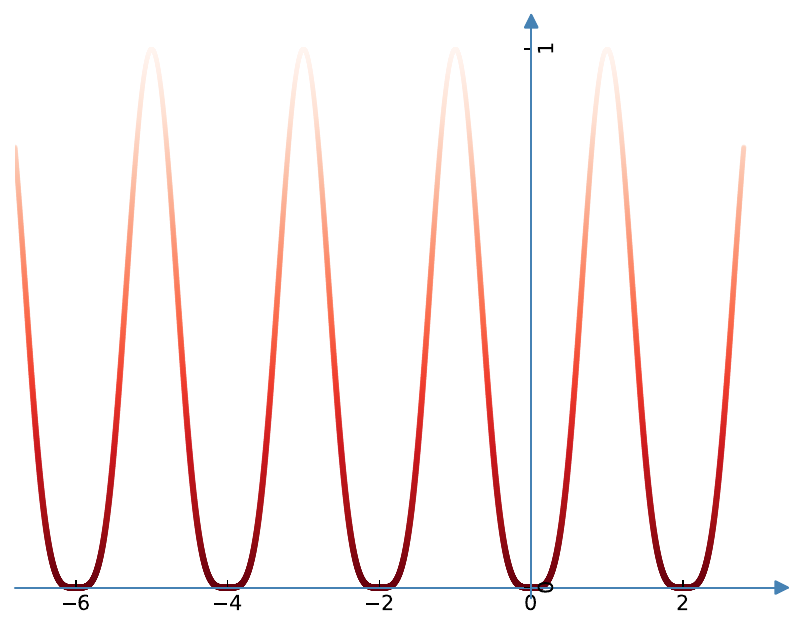}
\small
\caption{\small Integrity-well Loss.}
\label{fig:well}
\end{wrapfigure}

We refine the topology estimate by introducing a self-optimization mechanism, using an integrity-well loss on topological invariants. The integrity-well loss function is defined as
\begin{equation*}
w_{int}(x) = \left(\sin\left(\pi x - \frac{\pi}{2}\right) + 1\right)^2,
\end{equation*}
illustrated in Fig. \ref{fig:well}.
This loss is designed to favor integer values and even-numbered invariants (aligned with the Riemann-Hurwitz formula Eq.\eqref{Riemann-Hurwitz}), as most real-world 3D shapes have even-genus surfaces not exceeding \(2\).

The self-optimization mechanism backpropagates the integrity-well loss over local offsets, normals, area elements, and frames. In normal optimization, for example, the loss is added to the normal estimation loss function. Typically, the normal fields are initialized using negative gradients of the signed distance function (SDF) or local PCA. Given \(\mathbf{n}_i = (\varphi_i, \theta_i)\), the \(k\)-step self-optimization of normals minimizes the loss function for unit vector fields \({(\varphi_i, \theta_i)}_{i=1}^N\) as
\begin{align}\label{eq:gradient descent}
    &(\varphi_i, \theta_i)^{(k+1)} - (\varphi_i, \theta_i)^{(k)} \nonumber\\
    =& - lr \frac{\partial}{\partial (\varphi_i, \theta_i)}\left( \|\chi^{(k)} - \chi_{Gt}\| + w_{int}\left(\chi^{(k)}\right)\right),
\end{align}
where \(\chi^{(k)}\) is the current Euler characteristic estimate, \(lr\) is the learning rate, and \(\chi_{Gt}\) is the ground truth. If \(\chi_{Gt}\) is unavailable, global winding number supervision may be used as suggested in \cite{xu2023globally}.

\begin{wrapfigure}{l}{0.0cm}
\includegraphics[width=0.23\textwidth]{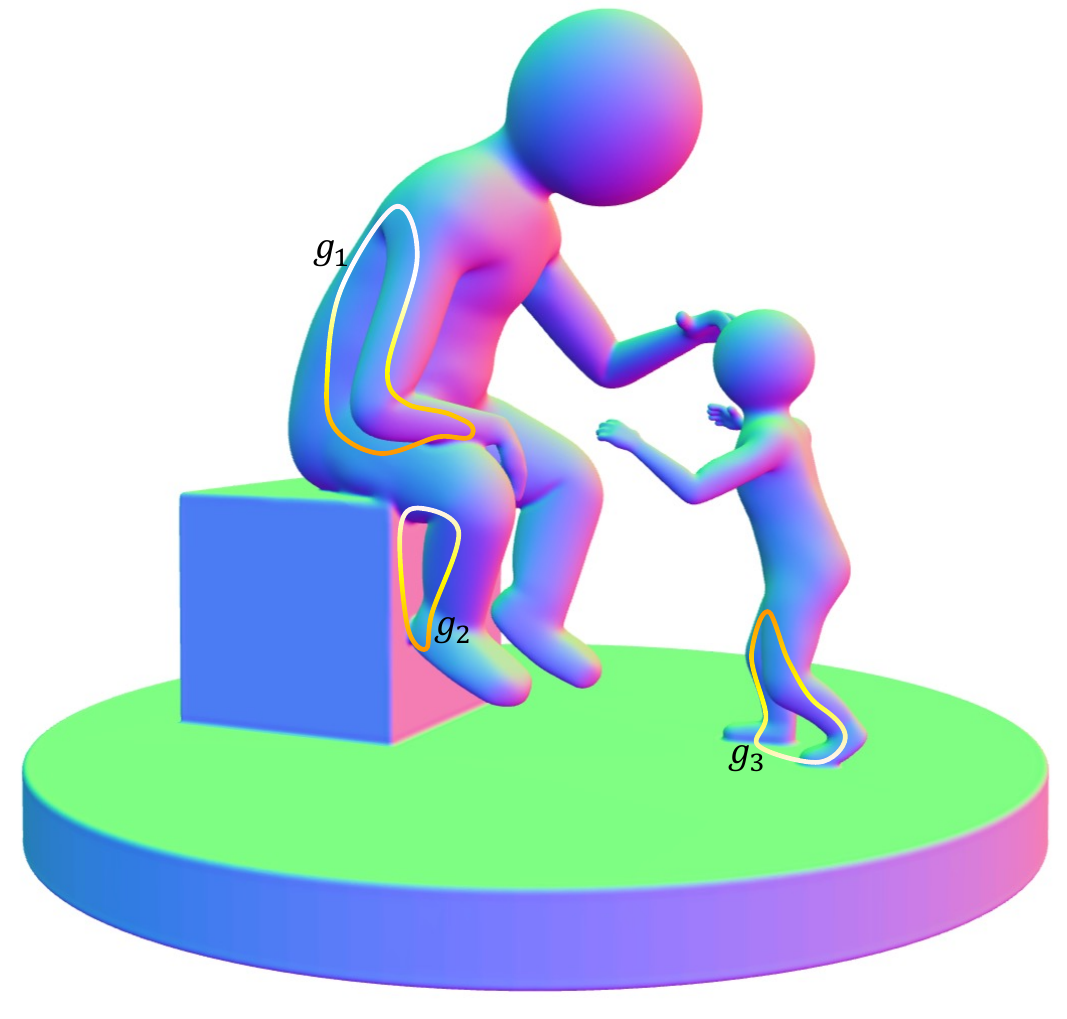}
\small
\caption{\small Three independent handles indicate an estimated genus of \(g(M) = 3\).}
\label{fig:genus}
\end{wrapfigure}

\begin{figure}[ht]
\centering
\includegraphics[width=0.5\textwidth]{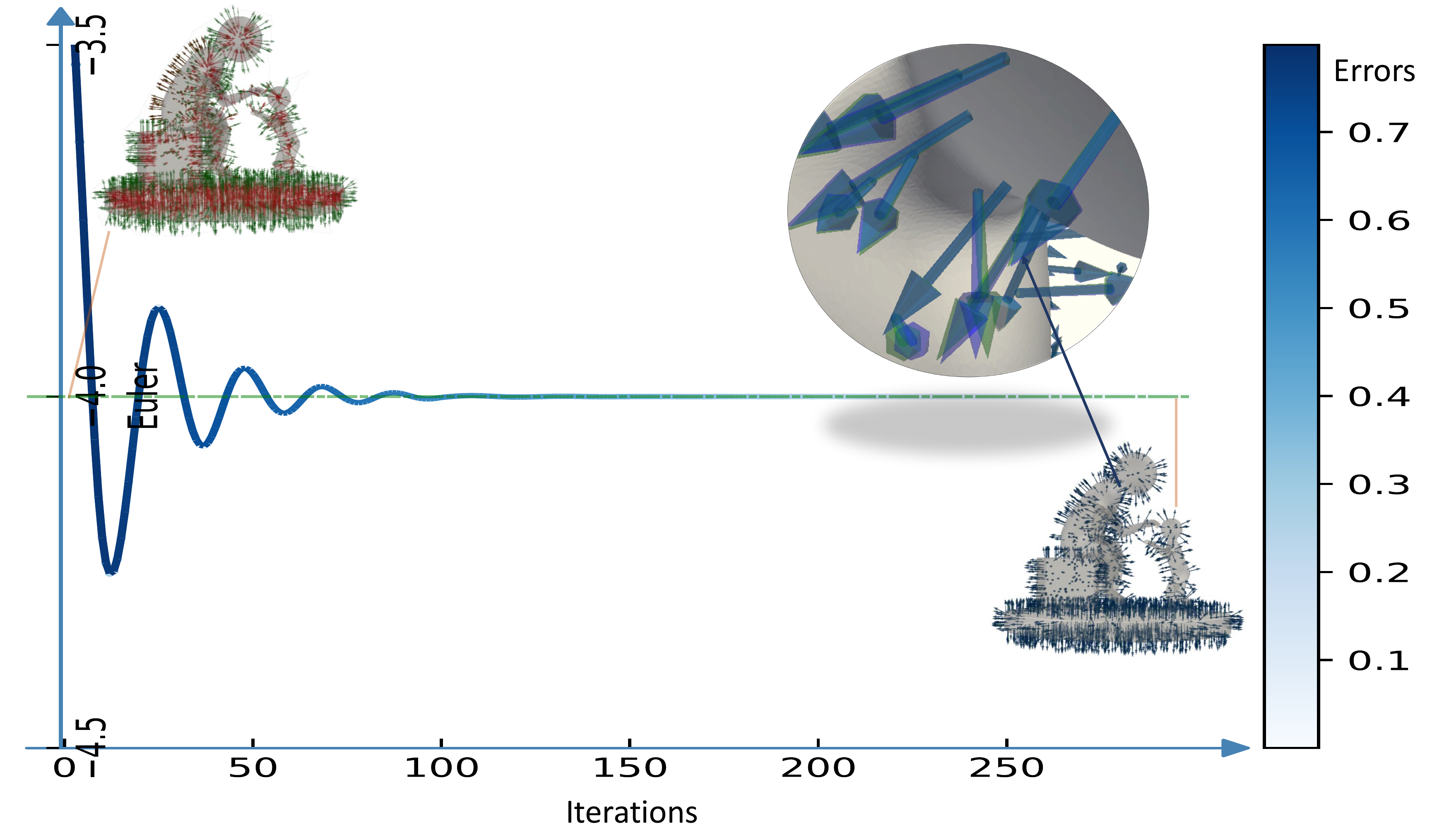}
\caption{\small Convergence of the self-optimization over normals.}
\label{fig:selfoptimization}
\end{figure}

Starting with ambiguous normal estimates, the optimization process refines these normals, leading to a near-integer approximation of the topology. For example, in our experiment with the "Father's Strength" model, the Euler characteristic converged to \(-4 \pm 0.02\) from an initial estimate of \(-3.5 \pm 0.02\), accurately indicating a genus \(g = 3\), consistent with the observed model structure shown in Fig.\ref{fig:genus}. Details of the optimization process are illustrated in Fig.\ref{fig:selfoptimization}.

With these novel algorithms and the designed architecture, we achieved a differentiable estimation of the global topology of 3D shapes. Algorithm \ref{alg:topologyestimation} summarizes the complete pipeline of our method. For other forms of 3D representations, point clouds can be easily extracted, and additional information about local frames can be naturally incorporated. For instance, gradient vectors of implicit field can imply normal directions, and Gaussian Splatting can directly provide local frames through the rotation component of the covariance matrix. Therefore, our method can be applied to various 3D representations.

In the following section, we will present experimental results that demonstrate our method's superior performance compared to existing approaches.

\begin{algorithm}[h!]
  \caption{Differentiable Topology Estimation}
   \hspace*{0.02in} {\bf Input:} 3D representation\\
   \hspace*{0.02in} {\bf Output:} Euler characteristic \(\chi_M\), genus \(g(M)\) \\
  \vspace{-0.4cm}
  \begin{algorithmic}[1]

    \STATE \textbf{Point cloud Extraction:} Sample point cloud from localized 3D representations.
      
    \STATE \textbf{Neighborhood Detection:} For each point \(\mathbf{p}_i \in \mathcal{P}\), find the \(k\)-nearest neighbors, denoted as \(\mathcal{N}_i\), to capture local geometric context.
    
    \STATE \textbf{Local Frame Construction:} For each point, apply local PCA to compute the moving orthonormal frame \(\{\mathbf{t}_i, \mathbf{t}^{\prime}_i, \mathbf{n}_i\}\).
    
    \STATE \textbf{Tangent Space Approximation:} Compute the local tangent coordinates \(\mathrm{d}\widetilde{\mathbf{p}}_{ij}\) for each neighbor in \(\mathcal{N}_i\) using Eq. \eqref{local tangent coordinate system}. Approximate the area element \(A_i\) with a Monte Carlo ratio, as per Eq. \eqref{area element}.
    
    \STATE \textbf{Weingarten Map Estimation:} For each \(\mathcal{N}_i\), project normal differences onto the tangent space to obtain \(\mathrm{d}\tilde{\mathbf{n}}_{ij}\), then estimate the Weingarten map \(W_i\). Use either Algorithm \ref{alg:robustestimationofW} or Eqs. \eqref{eq:pseudo-inverse} and \eqref{eq:symmetrizing}.
    
    \STATE \textbf{Gaussian Curvature Computation:} Calculate Gaussian curvature \(K_i\) using Eq. \eqref{Gaussian curvature} or its approximation in Eq. \eqref{approximate gaussian curvature}.
    
    \STATE \textbf{Initial Topology Estimation:} Estimate the initial Euler characteristic \(\chi^{(0)}_M\) by applying the discrete Gauss-Bonnet theorem, Eq. \eqref{eq:euler}.
    
    \STATE \textbf{Self-Optimization:} Refine location $\mathbf{p}_i$ and normal vectors \(\mathbf{n}_i\) with the gradient descent of the integrity-well loss.
    
    \STATE \textbf{Iterative Refinement:} Repeat Step 7 for \(K\) iterations, outputting the final Euler characteristic \(\chi_M = \chi^{(K)}_M\). Compute the genus \(g(M)\) using Eq. \eqref{Riemann-Hurwitz}.
  \end{algorithmic}
  \label{alg:topologyestimation}
\end{algorithm}

\section{Experiments}
\label{sec:experiments}
\begin{figure*}[t]
\centering
\includegraphics[width=0.9\textwidth]{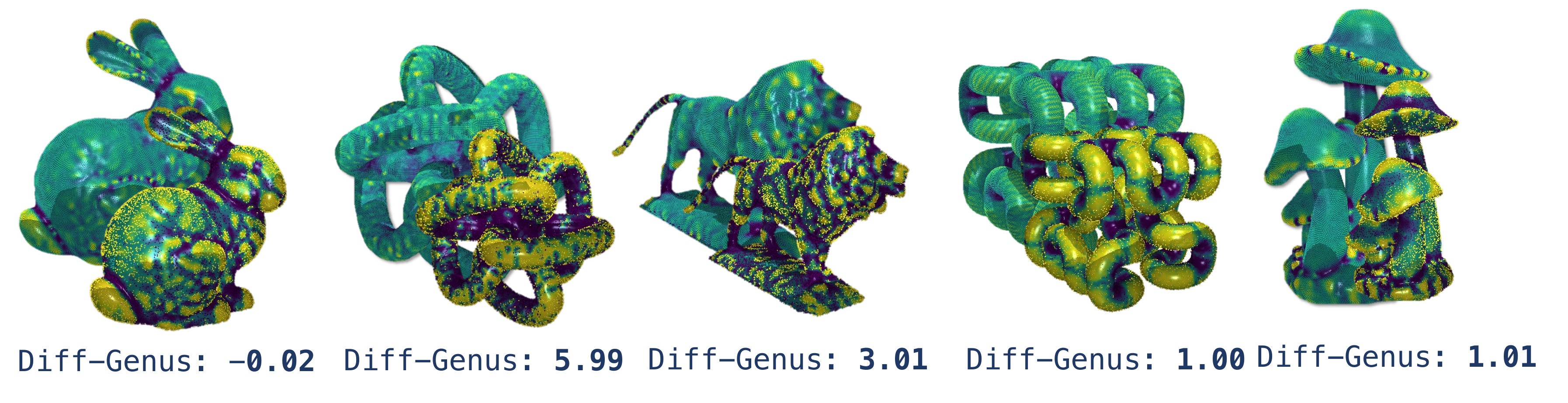}
\caption{Comparison on noised parameterized surface samples.}
\label{fig:Even}
\end{figure*}

In this section, we present the experimental results of our method to demonstrate its effectiveness and efficiency. Comparisons with existing methods show that our approach achieves superior performance across various 3D models for both curvature and topology estimation. The ablation study illustrates the impact of our proposed simplifications in the robust estimation of the Weingarten map and the self-optimization mechanism. Our method is implemented in PyTorch with CUDA (v2.0, CUDA 11.8) on Linux (Ubuntu 20.04). All experiments are conducted on a single NVIDIA RTX A6000 GPU with 48GB of memory and an AMD EPYC 7313P 16-core CPU.


\subsection{Performance of Curvature Estimation}
\begin{table*}[t]
\caption{Comparison of Curvature Estimation on Parametric Surface }
\label{tab:curvatureestimation}
\resizebox{1.0\linewidth}{!}{\begin{tabular}{|lllllllllllllllllll|}
\hline
& \multicolumn{9}{c|}{\textbf{Ellipsoid}}                & \multicolumn{9}{c|}{\textbf{Torus}}  \\ \hline
\multicolumn{1}{|l|}{}                 & \multicolumn{3}{c|}{$x^2+y^2+z^2=1$, 2k}                                           & \multicolumn{3}{c|}{$x^2 + \frac{y^2}{2}+\frac{y^2}{4}=1$,  10k}                 & \multicolumn{3}{c|}{10k, +2.5\%$\mathcal{N}$}                  & \multicolumn{3}{c|}{$R=5, r=1$, 10k}                      & \multicolumn{3}{c|}{$R=5, r=3$, 70k}              & \multicolumn{3}{c|}{70k, +2.5\%$\mathcal{N}$}                                \\ \hline
\multicolumn{19}{|c|}{Gaussian Curvature}               \\ \hline
\multicolumn{1}{|l|}{log-Error}            & Max             & \multicolumn{1}{l|}{Mean}            & \multicolumn{1}{l|}{Euler}           & Max             & \multicolumn{1}{l|}{Mean}            & \multicolumn{1}{l|}{Euler}           & Max            & \multicolumn{1}{l|}{Mean}            & \multicolumn{1}{l|}{Euler}           & Max             & \multicolumn{1}{l|}{Mean}            & \multicolumn{1}{l|}{Euler}           & Max             & \multicolumn{1}{l|}{Mean}            & \multicolumn{1}{l|}{Euler}            & Max             & Mean                                 & Euler           \\ \hline
\multicolumn{1}{|l|}{\textit{PCA}}     & 8.020           & \multicolumn{1}{l|}{7.450}           & \multicolumn{1}{l|}{-7.2e+3}         & 7.931           & \multicolumn{1}{l|}{7.425}           & \multicolumn{1}{l|}{-1.4e+4}         & 9.035          & \multicolumn{1}{l|}{8.391}           & \multicolumn{1}{l|}{-5.6e+4}         & 6.583           & \multicolumn{1}{l|}{6.317}           & \multicolumn{1}{l|}{-4.7e+4}         & 9.376           & \multicolumn{1}{l|}{8.886}           & \multicolumn{1}{l|}{-2.0e+6}          & 10.156          & \multicolumn{1}{l|}{9.592}           & -4.4e+6         \\
\multicolumn{1}{|l|}{\textit{Taubin}}  & 2.620           & \multicolumn{1}{l|}{-0.260}          & \multicolumn{1}{l|}{\textbf{2.149}}  & 3.079           & \multicolumn{1}{l|}{-0.129}          & \multicolumn{1}{l|}{3.125}           & 5.021          & \multicolumn{1}{l|}{2.521}           & \multicolumn{1}{l|}{8.268}           & 3.739           & \multicolumn{1}{l|}{0.617}           & \multicolumn{1}{l|}{16.41}           & 3.324           & \multicolumn{1}{l|}{-0.360}          & \multicolumn{1}{l|}{16.04}            & 4.271           & \multicolumn{1}{l|}{1.673}           & 77.48           \\
\multicolumn{1}{|l|}{\textit{Normal}}  & -0.010          & \multicolumn{1}{l|}{-0.040}          & \multicolumn{1}{l|}{0.006}           & 0.176           & \multicolumn{1}{l|}{-0.045}          & \multicolumn{1}{l|}{0.024}           & \textbf{0.249} & \multicolumn{1}{l|}{-0.045}          & \multicolumn{1}{l|}{0.159}           & -0.234          & \multicolumn{1}{l|}{-0.311}          & \multicolumn{1}{l|}{1.34}            & -0.640          & \multicolumn{1}{l|}{-0.830}          & \multicolumn{1}{l|}{\textbf{0.58}}    & -0.525          & \multicolumn{1}{l|}{-0.840}          & 4.94            \\
\multicolumn{1}{|l|}{\textit{Quadric}} & -1.980          & \multicolumn{1}{l|}{-2.540}          & \multicolumn{1}{l|}{\textbf{2.005}}  & \textbf{-0.295} & \multicolumn{1}{l|}{\textbf{-0.583}} & \multicolumn{1}{l|}{0.502}           & 3.790          & \multicolumn{1}{l|}{1.618}           & \multicolumn{1}{l|}{6.976}           & 2.775           & \multicolumn{1}{l|}{-0.234}          & \multicolumn{1}{l|}{16.05}           & 2.527           & \multicolumn{1}{l|}{-0.795}          & \multicolumn{1}{l|}{15.66}            & 3.899           & \multicolumn{1}{l|}{0.728}           & 54.28           \\
\multicolumn{1}{|l|}{\textit{Our}}     & \textbf{-3.020} & \multicolumn{1}{l|}{\textbf{-3.700}} & \multicolumn{1}{l|}{\textbf{2.001}}  & 0.744           & \multicolumn{1}{l|}{-0.074}          & \multicolumn{1}{l|}{\textbf{1.551}}  & 0.800          & \multicolumn{1}{l|}{\textbf{-0.081}} & \multicolumn{1}{l|}{\textbf{1.512}}  & \textbf{-0.437} & \multicolumn{1}{l|}{\textbf{-0.893}} & \multicolumn{1}{l|}{\textbf{-0.04}}  & \textbf{-2.538} & \multicolumn{1}{l|}{\textbf{-3.523}} & \multicolumn{1}{l|}{\textbf{-0.0002}} & \textbf{-1.038} & \multicolumn{1}{l|}{\textbf{-2.420}} & \textbf{-0.016} \\ \hline
\multicolumn{19}{|c|}{Mean Curvature}                                                                                                     \\ \hline
log-Error                                  & Max             & \multicolumn{1}{l|}{Mean}            & \multicolumn{1}{l|}{Time}            & Max             & \multicolumn{1}{l|}{Mean}            & \multicolumn{1}{l|}{Time}            & Max            & \multicolumn{1}{l|}{Mean}            & \multicolumn{1}{l|}{Time}            & Max             & \multicolumn{1}{l|}{Mean}            & \multicolumn{1}{l|}{Time}            & Max             & \multicolumn{1}{l|}{Mean}            & \multicolumn{1}{l|}{Time}             & Max             & Mean                                 & Time            \\ \hline
\multicolumn{1}{|l|}{\textit{PCA}}     & 4.011           & \multicolumn{1}{l|}{3.680}           & \multicolumn{1}{l|}{0.673}           & 3.984           & \multicolumn{1}{l|}{3.674}           & \multicolumn{1}{l|}{0.913}           & 4.522          & \multicolumn{1}{l|}{4.109}           & \multicolumn{1}{l|}{0.702}           & 3.334           & \multicolumn{1}{l|}{3.181}           & \multicolumn{1}{l|}{0.396}           & 4.688           & \multicolumn{1}{l|}{4.407}           & \multicolumn{1}{l|}{1.881}            & 5.079           & \multicolumn{1}{l|}{4.744}           & 1.834           \\
\multicolumn{1}{|l|}{\textit{Taubin}}  & 1.247           & \multicolumn{1}{l|}{-0.845}          & \multicolumn{1}{l|}{0.458}           & 1.555           & \multicolumn{1}{l|}{-0.505}          & \multicolumn{1}{l|}{0.302}           & 2.279          & \multicolumn{1}{l|}{0.888}           & \multicolumn{1}{l|}{2.279}           & 1.628           & \multicolumn{1}{l|}{-0.144}          & \multicolumn{1}{l|}{-0.136}          & 1.469           & \multicolumn{1}{l|}{-0.692}          & \multicolumn{1}{l|}{1.205}            & 1.915           & \multicolumn{1}{l|}{0.448}           & 1.008           \\
\multicolumn{1}{|l|}{\textit{Normal}}  & -0.001          & \multicolumn{1}{l|}{-0.001}          & \multicolumn{1}{l|}{0.451}           & 0.301           & \multicolumn{1}{l|}{-0.096}          & \multicolumn{1}{l|}{0.323}           & \textbf{0.353} & \multicolumn{1}{l|}{-0.114}          & \multicolumn{1}{l|}{0.394}           & -0.005          & \multicolumn{1}{l|}{\textbf{-0.876}} & \multicolumn{1}{l|}{-0.112}          & \textbf{-0.304} & \multicolumn{1}{l|}{\textbf{-1.272}} & \multicolumn{1}{l|}{1.038}            & -0.101          & \multicolumn{1}{l|}{\textbf{-1.037}} & 1.011           \\
\multicolumn{1}{|l|}{\textit{Quadric}} & -2.284          & \multicolumn{1}{l|}{-2.854}          & \multicolumn{1}{l|}{0.475}           & -0.299 & \multicolumn{1}{l|}{-0.580} & \multicolumn{1}{l|}{0.394}           & 2.467          & \multicolumn{1}{l|}{0.575}           & \multicolumn{1}{l|}{0.433}           & 1.373           & \multicolumn{1}{l|}{-0.280}          & \multicolumn{1}{l|}{-0.013}          & 1.004           & \multicolumn{1}{l|}{-0.807}          & \multicolumn{1}{l|}{1.096}            & 2.522           & \multicolumn{1}{l|}{0.093}           & 1.066           \\
\multicolumn{1}{|l|}{\textit{Our}}     & \textbf{-3.146} & \multicolumn{1}{l|}{\textbf{-3.916}} & \multicolumn{1}{l|}{\textbf{-3.042}} & \textbf{-0.387}           & \multicolumn{1}{l|}{\textbf{-1.764}}          & \multicolumn{1}{l|}{\textbf{-3.013}} & \textbf{0.364} & \multicolumn{1}{l|}{\textbf{-0.716}} & \multicolumn{1}{l|}{\textbf{-3.017}} & \textbf{-0.169} & \multicolumn{1}{l|}{-0.336}          & \multicolumn{1}{l|}{\textbf{-3.410}} & \textbf{-0.339} & \multicolumn{1}{l|}{-0.538}          & \multicolumn{1}{l|}{\textbf{-2.439}}  & \textbf{-0.297} & \multicolumn{1}{l|}{-0.538}          & \textbf{-2.434} \\ \hline

\end{tabular}}

\end{table*}
We first evaluate the performance of the self-adjoint Weingarten map estimation method on the task of curvature estimation, comparing it with state-of-the-art methods, including the Taubin method \cite{taubin1995estimating}, the robust statistical method \cite{kalogerakis2007robust}, and the quadratic fitting method \cite{wardetzky2007discrete}. Since the theoretical curvature values can only be obtained from analytical surfaces, ground truth supervised evaluations are conducted solely on ellipsoids and tori with varying parameters, sampling densities, and noise levels. Estimation results for both Gaussian and mean curvatures are assessed by the maximum absolute error (Max), mean absolute error (Mean) over the entire surface, and the Euler characteristic (Euler) to measure error accumulation. The theoretical values of Gaussian and mean curvature are derived directly from the definitions of curvature in differential geometry.

\begin{table*}[ht]

\caption{Comparison of Topology Estimation on 3D Models }
\label{tab:topologyestimation}
\resizebox{1.0\linewidth}{!}{
\begin{tabular}{|l|ll|llll|llll|llll|}
\hline
                       & \multicolumn{2}{c|}{Ground-Truth}                    & \multicolumn{4}{c|}{5k}                                                 & \multicolumn{4}{c|}{10k}                                                 & \multicolumn{4}{c|}{10k + 25\%}                                         \\ \hline
\textit{3D Model Name} & \multicolumn{1}{l|}{\textit{Euler}} & \textit{Genus} & \textit{Taubin} & \textit{Normal} & \textit{Quadric} & \textit{Ours}    & \textit{Taubin} & \textit{Normal} & \textit{Quadric} & \textit{Ours}     & \textit{Taubin} & \textit{Normal} & \textit{Quadric} & \textit{Ours}    \\ \hline
RindStrips             & \multicolumn{1}{l|}{-12}            & 7              & -187.574        & 42.150          & 24.451           & \textbf{-12.721} & -364.580        & 18.763          & 31.421           & \textbf{-12.281}  & -513.549        & 24.296          & 34.491           & \textbf{-11.316} \\
ChicagoLion            & \multicolumn{1}{l|}{-4}             & 3              & -134.305        & 11.153          & -39.028          & -2.243           & -231.445        & 4.101           & 4.344            & \textbf{-3.891}   & -261.315        & 4.984           & 22.265           & -7.979           \\
HolesSculpture         & \multicolumn{1}{l|}{-402}           & 202            & -255.650        & 0.000           & 8.211            & -380.564         & -338.778        & 0.000           & 14.388           & \textbf{-403.540} & -211.279        & 0.000           & -1.280           & -195.819         \\
TriakisTetrahedron     & \multicolumn{1}{l|}{-20}            & 11             & -240.753        & 246.402         & -4.811           & -13.386          & -156.516        & 83.043          & \textbf{-18.717} & \textbf{-19.301}  & -278.019        & 109.510         & -13.049          & -27.020          \\
OrganicSphere          & \multicolumn{1}{l|}{2}              & 0              & -43.351         & 50.699          & 4.538            & \textbf{1.911}   & -44.778         & 14.752          & 5.289            & \textbf{2.015}    & -168.501        & 17.065          & 49.377           & \textbf{1.417}   \\
Hilb64Thick            & \multicolumn{1}{l|}{0}              & 1              & -97.563         & 904.153         & 11.984           & 1.641            & -168.468        & 271.507         & 10.889           & \textbf{0.707}    & -277.612        & 290.248         & 5.732            & 2.586            \\
Quirrel                & \multicolumn{1}{l|}{2}              & 0              & -33.761         & 200.322         & 3.842            & \textbf{1.911}   & -27.654         & 52.086          & 4.752            & \textbf{2.293}    & -76.149         & 74.805          & 4.702            & \textbf{1.348}   \\
Icosphere              & \multicolumn{1}{l|}{2}              & 0              & 0.802           & 0.000           & \textbf{2.006}   & \textbf{1.985}   & \textbf{1.038}  & 0.000           & \textbf{1.967}   & \textbf{1.997}    & -332.351        & 0.003           & 3.132            & \textbf{1.952}   \\
Knot                   & \multicolumn{1}{l|}{0}              & 1              & -38.954         & 14.771          & \textbf{-0.928}  & \textbf{-0.854}  & -34.389         & 3.175           & \textbf{-0.895}  & \textbf{-0.376}   & -96.946         & 6.737           & -1.898           & \textbf{0.125}   \\
Venus                  & \multicolumn{1}{l|}{2}              & 0              & -119.564        & 2694.091        & 5.868            & \textbf{1.534}   & -110.949        & 861.677         & \textbf{2.980}   & \textbf{2.100}    & -157.928        & 911.814         & 4.339            & \textbf{1.471}   \\
Lcositetrahedron       & \multicolumn{1}{l|}{-44}            & 23             & -166.592        & 2102.811        & 6.868            & \textbf{-42.751} & -229.629        & 698.427         & -16.535          & \textbf{-43.262}  & -256.594        & 700.772         & -17.060          & -35.146          \\
Kitten                 & \multicolumn{1}{l|}{0}              & 1              & -38.744         & 0.000           & 0.530            & \textbf{0.515}   & -41.165         & 0.000           & 1.027            & \textbf{0.504}    & -334.060        & 0.001           & 0.659            & \textbf{-0.113}  \\
TeaPot                 & \multicolumn{1}{l|}{0}              & 1              & -43.676         & 4507.909        & 6.331            & \textbf{-0.473}  & -60.219         & 2243.370        & 5.132            & -1.136            & -50.036         & 1586.369        & 4.848            & -2.123           \\
Bunny                  & \multicolumn{1}{l|}{2}              & 0              & -62.907         & 2181.226        & 8.352            & \textbf{1.897}   & -128.493        & 836.904         & 3.647            & \textbf{1.898}    & -109.532        & 728.377         & 14.158           & \textbf{1.370}   \\
FatherStrength         & \multicolumn{1}{l|}{-4}             & 3              & -77.286         & 0.001           & 5.762            & -2.613           & -107.531        & 0.000           & 11.583           & \textbf{-3.671}   & -345.582        & 0.003           & 3.190            & \textbf{-3.626}  \\
QueenAnneChair         & \multicolumn{1}{l|}{-2}             & 2              & -90.733         & 73.852          & 0.369            & \textbf{-1.214}  & -98.113         & 23.431          & -5.794           & \textbf{-2.324}   & -263.281        & 31.454          & \textbf{-2.391}  & \textbf{-1.519}  \\
Art3dprint             & \multicolumn{1}{l|}{-102}           & 52             & -246.946        & 0.000           & -1.416           & -90.612          & -329.704        & 0.000           & -17.058          & -90.593           & -230.682        & 0.000           & 3.206            & -109.380         \\
Art3dprint2            & \multicolumn{1}{l|}{-18}            & 10             & -170.480        & 0.000           & -8.640           & \textbf{-17.703 }         & -282.578        & 0.000           & -14.653          & \textbf{-17.545}           & -237.336        & 0.000           & 5.192            & -12.158          \\
Mushroom               & \multicolumn{1}{l|}{0}              & 1              & -106.634        & 2468.301        & -2.093           & \textbf{-0.018}  & -98.255         & 733.907         & 2.477            & \textbf{0.429}    & -135.657        & 654.618         & 6.820            & -2.199           \\
Genus6surface          & \multicolumn{1}{l|}{-10}            & 6              & -195.469        & 38.372          & -7.271           & \textbf{-9.204 }          & -117.047        & 11.750          & -11.734          & \textbf{-10.072}  & -284.386        & 13.549          & \textbf{-9.051}  & \textbf{-9.626}  \\ \hline
\end{tabular}}
\end{table*}

Specifically, for an ellipsoid $\frac{x^2}{a^2}+\frac{y^2}{b^2}+\frac{z^2}{c^2}=1$, we have
\begin{equation}
\begin{aligned}
& K_g(x,y,z) = \frac{1}{a^2b^2c^2\left(\frac{x^2}{a^4}+\frac{y^2}{b^4}+\frac{z^2}{c^4}\right)^2},\\
& H(x,y,z) = \frac{x^2+y^2+z^2-a^2-b^2-c^2}{2(a^2b^2c^2)\sqrt{\left(\frac{x^2}{a^4}+\frac{y^2}{b^4}+\frac{z^2}{c^4}\right)^3}}.
\end{aligned}
\end{equation}
For a torus parameterized by
\begin{equation}
\left\{  
\begin{array}{l l}
x = (R+r\cos(u))\cos(v),\\
y = (R+r\cos(u))\sin(v),\\
z = r\sin(u),
\end{array}
\right.
\end{equation}
we obtain the curvatures as
\begin{equation}
\begin{aligned}
& K_g = \frac{\cos(v)}{r(R+r\cos(u))},\\
& H = \frac{1}{2}\left(\frac{1}{r}+\frac{\cos(v)}{R+r\cos(u)}\right).
\end{aligned}
\end{equation}

The results are presented in Table \ref{tab:curvatureestimation}, with errors recorded on a logarithmic scale. The comparison shows that our method outperforms state-of-the-art methods for both Gaussian and mean curvature estimation across different sampling densities and noise levels. Notably, our method uniquely achieves Euler characteristic estimation and resists error accumulation, showing insensitivity to noise. Additionally, by leveraging parallel computation, our method is significantly faster than traditional approaches, particularly for high-density point clouds or high-resolution implicit fields.

For general 3D models, where ground truth curvature is difficult to obtain, we indirectly evaluate curvature estimation via topology estimation. Additionally, curvature-aware decimation of point clouds and surface reconstruction will further demonstrate the effectiveness of our curvature estimation, as discussed in the following subsections.

\subsection{Performance on Topology Estimation}

First, we evaluate the performance of our method on the topology estimation task, comparing it with the curvature-based methods introduced in the previous subsection. This comparison is conducted on 20 3D models with varying topologies and geometric complexities from the SHREC dataset \cite{li2015comparison}, the ModelNet40 dataset \cite{wu20153d}, and the ShapeNet dataset \cite{chang2015shapenet}. The results, presented in Table \ref{tab:topologyestimation}, demonstrate that our method achieves an average accuracy of nearly 90\% for topology estimation, significantly outperforming state-of-the-art methods.

\begin{figure}[ht!]
\centering
\includegraphics[width=0.46\textwidth]{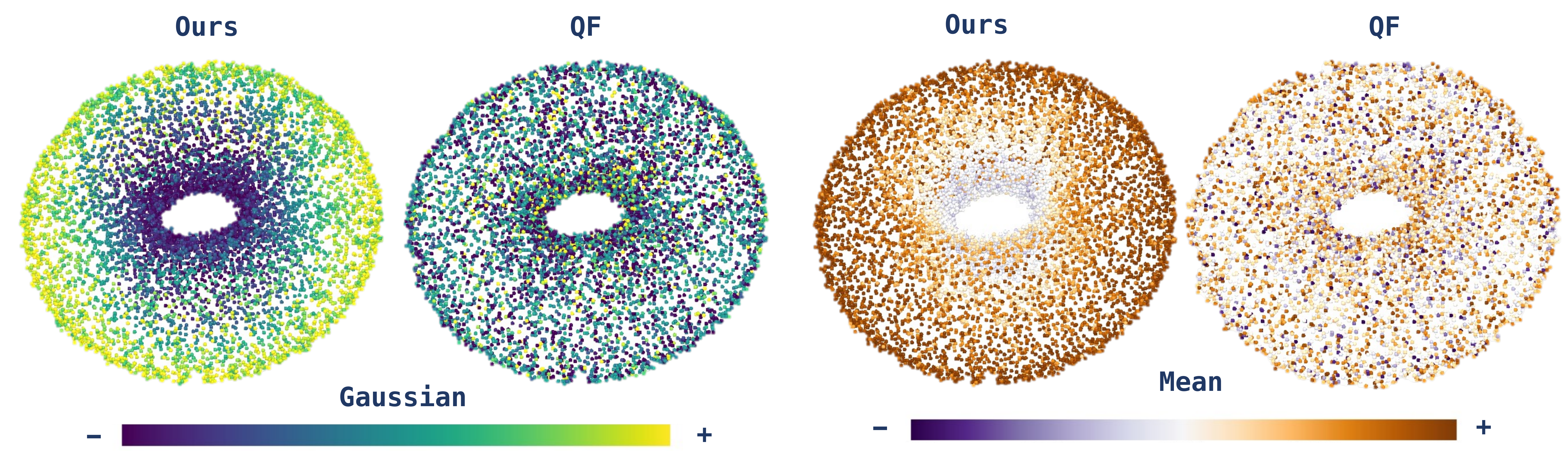}
\caption{Comparison on parameterized surface samples with noise.}
\label{fig:ParaSurf}
\end{figure}

Additionally, our experiments indicate that our method is robust to different sampling schemes and noise density. Uniform sampling generally poses greater implementation challenges as it requires careful attention to local surface areas. Through self-optimization during integration, our method obtains correct topology estimations with random sampling and added noise, as illustrated in Fig. \ref{fig:ParaSurf} and the final column in Table \ref{tab:topologyestimation}.

On the other hand, we compare the efficiency of our method with the topological data analysis (TDA) method, specifically persistent homology \cite{edelsbrunner2013persistent, zomorodian2004computing, otter2017roadmap}, which is the prevailing approach for topology estimation. Our comparison shows that our method is not only significantly faster than the TDA method but also less ambiguous.
Persistent homology (PH) constructs a 2D Vietoris-Rips complex \cite{hausmann1994vietoris} and generates a persistence diagram that represents the topological features of a shape. This diagram captures information about the $0$-dimensional and $1$-dimensional homology groups: the $0$-dimensional group corresponds to the number of connected components, while the $1$-dimensional group reveals the genus (the number of holes). In PH, each segment on the resulting barcode corresponds to an independent topological feature, with a long-lasting $0$-dimensional bar indicating a stable connected component and a persistent $1$-dimensional bar signifying a stable hole. 

\begin{figure}[h!]
\centering
\includegraphics[width=0.45\textwidth]{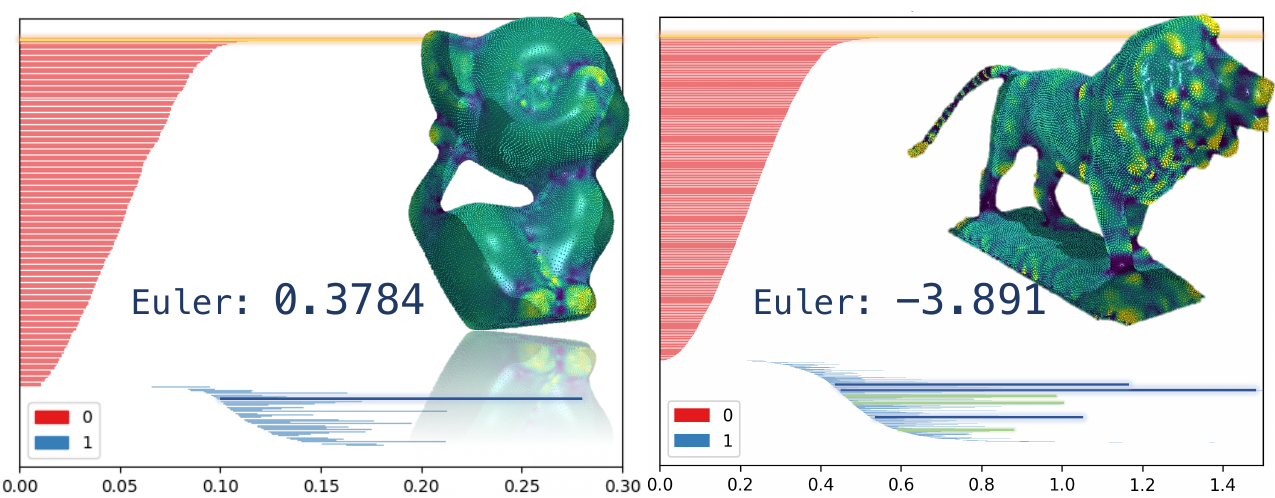}
\caption{Barcode of Persistence diagram}
\label{fig:barcode}
\end{figure}

Interpreting topology from these barcodes, however, can be challenging due to scaling differences across different genera.
For instance, Fig. \ref{fig:barcode} shows the persistence diagrams' barcodes for the topology estimation of the Kitten (connected, 1-genus) and ChicagoLion (connected, 3-genus). Both models exhibit a single, long-standing $0$-bar. While the Kitten has a unique long-lasting $1$-bar, the ChicagoLion shows more than three components, which makes it hard to align with its known topological structures. Additionally, PH is computationally intensive; in our experiments, PH required 49.86 seconds to process a 7k-point cloud. In contrast, our method directly approximates the genus with minimal ambiguity, which is effectively resolved through a self-optimization mechanism. Furthermore, unlike the iterative loops in PH, which are difficult to parallelize, our approach is fully differentiable and can be efficiently implemented on GPUs.

\section{Conclusion}

In this paper, we propose a differentiable algorithm to estimate the curvature and global topology of 3D shapes. Curvature estimation provides an important intrinsic geometric feature of 3D shapes. Our instant topology estimation, achieved for the first time, provides a reliable initialization for mesh-deformation-based reconstruction and generation tasks. Differentibility allows our method to be integrated into deep learning frameworks, enabling end-to-end training of neural networks for shape analysis tasks. We demonstrate the effectiveness of our method on both synthetic and real-world datasets.
\bibliographystyle{IEEEtran}
\bibliography{main}

\begin{thebibliography}{10}
\providecommand{\url}[1]{#1}
\csname url@samestyle\endcsname
\providecommand{\newblock}{\relax}
\providecommand{\bibinfo}[2]{#2}
\providecommand{\BIBentrySTDinterwordspacing}{\spaceskip=0pt\relax}
\providecommand{\BIBentryALTinterwordstretchfactor}{4}
\providecommand{\BIBentryALTinterwordspacing}{\spaceskip=\fontdimen2\font plus
\BIBentryALTinterwordstretchfactor\fontdimen3\font minus \fontdimen4\font\relax}
\providecommand{\BIBforeignlanguage}[2]{{%
\expandafter\ifx\csname l@#1\endcsname\relax
\typeout{** WARNING: IEEEtran.bst: No hyphenation pattern has been}%
\typeout{** loaded for the language `#1'. Using the pattern for}%
\typeout{** the default language instead.}%
\else
\language=\csname l@#1\endcsname
\fi
#2}}
\providecommand{\BIBdecl}{\relax}
\BIBdecl

\bibitem{genova2019learning}
K.~Genova, F.~Cole, D.~Vlasic, A.~Sarna, W.~T. Freeman, and T.~Funkhouser, ``Learning shape templates with structured implicit functions,'' in \emph{Proceedings of the IEEE/CVF International Conference on Computer Vision}, 2019, pp. 7154--7164.

\bibitem{park2019deepsdf}
J.~J. Park, P.~Florence, J.~Straub, R.~Newcombe, and S.~Lovegrove, ``Deepsdf: Learning continuous signed distance functions for shape representation,'' in \emph{Proceedings of the IEEE/CVF conference on computer vision and pattern recognition}, 2019, pp. 165--174.

\bibitem{marschner2023constructive}
Z.~Marschner, S.~Sell{\'a}n, H.-T.~D. Liu, and A.~Jacobson, ``Constructive solid geometry on neural signed distance fields,'' in \emph{SIGGRAPH Asia 2023 Conference Papers}, 2023, pp. 1--12.

\bibitem{zhang2021learning}
J.~Zhang, Y.~Yao, and L.~Quan, ``Learning signed distance field for multi-view surface reconstruction,'' in \emph{Proceedings of the IEEE/CVF International Conference on Computer Vision}, 2021, pp. 6525--6534.

\bibitem{mildenhall2021nerf}
B.~Mildenhall, P.~P. Srinivasan, M.~Tancik, J.~T. Barron, R.~Ramamoorthi, and R.~Ng, ``Nerf: Representing scenes as neural radiance fields for view synthesis,'' \emph{Communications of the ACM}, vol.~65, no.~1, pp. 99--106, 2021.

\bibitem{muller2022instant}
T.~M{\"u}ller, A.~Evans, C.~Schied, and A.~Keller, ``Instant neural graphics primitives with a multiresolution hash encoding,'' \emph{ACM transactions on graphics (TOG)}, vol.~41, no.~4, pp. 1--15, 2022.

\bibitem{poole2022dreamfusion}
B.~Poole, A.~Jain, J.~T. Barron, and B.~Mildenhall, ``Dreamfusion: Text-to-3d using 2d diffusion,'' \emph{arXiv preprint arXiv:2209.14988}, 2022.

\bibitem{kerbl20233d}
B.~Kerbl, G.~Kopanas, T.~Leimk{\"u}hler, and G.~Drettakis, ``3d gaussian splatting for real-time radiance field rendering.'' \emph{ACM Trans. Graph.}, vol.~42, no.~4, pp. 139--1, 2023.

\bibitem{wuc2024recent}
T.~Wu, Y.-J. Yuan, L.-X. Zhang, J.~Yang, Y.-P. Cao, L.-Q. Yan, and L.~Gao, ``Recent advances in 3d gaussian splatting,'' \emph{Computational Visual Media}, vol.~10, no.~4, pp. 613--642, 2024.

\bibitem{chen2024survey}
G.~Chen and W.~Wang, ``A survey on 3d gaussian splatting,'' \emph{arXiv preprint arXiv:2401.03890}, 2024.

\bibitem{huang20242d}
B.~Huang, Z.~Yu, A.~Chen, A.~Geiger, and S.~Gao, ``2d gaussian splatting for geometrically accurate radiance fields,'' in \emph{ACM SIGGRAPH 2024 Conference Papers}, 2024, pp. 1--11.

\bibitem{pascucci2007robust}
V.~Pascucci, G.~Scorzelli, P.-T. Bremer, and A.~Mascarenhas, ``Robust on-line computation of reeb graphs: simplicity and speed,'' in \emph{ACM SIGGRAPH 2007 papers}, 2007, pp. 58--es.

\bibitem{milnor1963morse}
J.~W. Milnor, \emph{Morse theory}.\hskip 1em plus 0.5em minus 0.4em\relax Princeton university press, 1963, no.~51.

\bibitem{edelsbrunner2008computational}
H.~Edelsbrunner and J.~Harer, ``Computational topology,'' \emph{Duke University}, 2008.

\bibitem{edelsbrunner2002topological}
Edelsbrunner, Letscher, and Zomorodian, ``Topological persistence and simplification,'' \emph{Discrete \& computational geometry}, vol.~28, pp. 511--533, 2002.

\bibitem{carlsson2009topology}
G.~Carlsson, ``Topology and data,'' \emph{Bulletin of the American Mathematical Society}, vol.~46, no.~2, pp. 255--308, 2009.

\bibitem{wu2008historical}
H.-H. Wu, ``Historical development of the gauss-bonnet theorem,'' \emph{Science in China Series A: Mathematics}, vol.~51, no.~4, p. 777, 2008.

\bibitem{merigot2010voronoi}
Q.~M{\'e}rigot, M.~Ovsjanikov, and L.~J. Guibas, ``Voronoi-based curvature and feature estimation from point clouds,'' \emph{IEEE Transactions on Visualization and Computer Graphics}, vol.~17, no.~6, pp. 743--756, 2010.

\bibitem{wardetzky2007discrete}
M.~Wardetzky, M.~Bergou, D.~Harmon, D.~Zorin, and E.~Grinspun, ``Discrete quadratic curvature energies,'' \emph{Computer Aided Geometric Design}, vol.~24, no. 8-9, pp. 499--518, 2007.

\bibitem{panozzo2010efficient}
D.~Panozzo, E.~Puppo, and L.~Rocca, ``Efficient multi-scale curvature and crease estimation,'' in \emph{2nd International Workshop on Computer Graphics, Computer Vision and Mathematics, GraVisMa}, 2010, pp. 9--16.

\bibitem{taubin1995estimating}
G.~Taubin, ``Estimating the tensor of curvature of a surface from a polyhedral approximation,'' in \emph{Proceedings of IEEE International Conference on Computer Vision}.\hskip 1em plus 0.5em minus 0.4em\relax IEEE, 1995, pp. 902--907.

\bibitem{cao2021efficient}
Y.~Cao, D.~Li, H.~Sun, A.~H. Assadi, and S.~Zhang, ``Efficient weingarten map and curvature estimation on manifolds,'' \emph{Machine Learning}, vol. 110, no.~6, pp. 1319--1344, 2021.

\bibitem{berkmann1994computation}
J.~Berkmann and T.~Caelli, ``Computation of surface geometry and segmentation using covariance techniques,'' \emph{IEEE Transactions on Pattern Analysis and Machine Intelligence}, vol.~16, no.~11, pp. 1114--1116, 1994.

\bibitem{kalogerakis2007robust}
E.~Kalogerakis, P.~Simari, D.~Nowrouzezahrai, and K.~Singh, ``Robust statistical estimation of curvature on discretized surfaces,'' in \emph{Symposium on Geometry Processing}, vol.~13, 2007, pp. 110--114.

\bibitem{yang2007direct}
P.~Yang and X.~Qian, ``Direct computing of surface curvatures for point-set surfaces.'' in \emph{PBG@ Eurographics}.\hskip 1em plus 0.5em minus 0.4em\relax Citeseer, 2007, pp. 29--36.

\bibitem{amenta2004domain}
N.~Amenta and Y.~J. Kil, ``The domain of a point set surfaces,'' in \emph{Eurographics Symposium on Point-based Graphics}, vol.~1, 2004.

\bibitem{aurenhammer1991voronoi}
F.~Aurenhammer, ``Voronoi diagrams—a survey of a fundamental geometric data structure,'' \emph{ACM Computing Surveys (CSUR)}, vol.~23, no.~3, pp. 345--405, 1991.

\bibitem{xu2023globally}
R.~Xu, Z.~Dou, N.~Wang, S.~Xin, S.~Chen, M.~Jiang, X.~Guo, W.~Wang, and C.~Tu, ``Globally consistent normal orientation for point clouds by regularizing the winding-number field,'' \emph{ACM Transactions on Graphics (TOG)}, vol.~42, no.~4, pp. 1--15, 2023.

\bibitem{chen1999lectures}
W.~Chen, S.-s. Chern, and K.~S. Lam, \emph{Lectures on differential geometry}.\hskip 1em plus 0.5em minus 0.4em\relax World Scientific Publishing Company, 1999, vol.~1.

\bibitem{ward1991general}
A.~Ward, ``A general analysis of sylvester's matrix equation,'' \emph{International Journal of Mathematical Education in Science and Technology}, vol.~22, no.~4, pp. 615--620, 1991.

\bibitem{luo2021geometric}
Y.~Luo, S.~Zhang, Y.~Cao, and H.~Sun, ``Geometric characteristics of the wasserstein metric on spd (n) and its applications on data processing,'' \emph{Entropy}, vol.~23, no.~9, p. 1214, 2021.

\bibitem{zhang2019new}
S.~Zhang, Y.~Cao, W.~Li, F.~Yan, Y.~Luo, and H.~Sun, ``A new riemannian structure in spd (n),'' in \emph{2019 IEEE International Conference on Signal, Information and Data Processing (ICSIDP)}.\hskip 1em plus 0.5em minus 0.4em\relax IEEE, 2019, pp. 1--5.

\bibitem{li2015comparison}
B.~Li, Y.~Lu, C.~Li, A.~Godil, T.~Schreck, M.~Aono, M.~Burtscher, Q.~Chen, N.~K. Chowdhury, B.~Fang \emph{et~al.}, ``A comparison of 3d shape retrieval methods based on a large-scale benchmark supporting multimodal queries,'' \emph{Computer Vision and Image Understanding}, vol. 131, pp. 1--27, 2015.

\bibitem{wu20153d}
Z.~Wu, S.~Song, A.~Khosla, F.~Yu, L.~Zhang, X.~Tang, and J.~Xiao, ``3d shapenets: A deep representation for volumetric shapes,'' in \emph{Proceedings of the IEEE conference on computer vision and pattern recognition}, 2015, pp. 1912--1920.

\bibitem{chang2015shapenet}
A.~X. Chang, T.~Funkhouser, L.~Guibas, P.~Hanrahan, Q.~Huang, Z.~Li, S.~Savarese, M.~Savva, S.~Song, H.~Su \emph{et~al.}, ``Shapenet: An information-rich 3d model repository,'' \emph{arXiv preprint arXiv:1512.03012}, 2015.

\bibitem{edelsbrunner2013persistent}
H.~Edelsbrunner, ``Persistent homology: theory and practice,'' 2013.

\bibitem{zomorodian2004computing}
A.~Zomorodian and G.~Carlsson, ``Computing persistent homology,'' in \emph{Proceedings of the twentieth annual symposium on Computational geometry}, 2004, pp. 347--356.

\bibitem{otter2017roadmap}
N.~Otter, M.~A. Porter, U.~Tillmann, P.~Grindrod, and H.~A. Harrington, ``A roadmap for the computation of persistent homology,'' \emph{EPJ Data Science}, vol.~6, pp. 1--38, 2017.

\bibitem{hausmann1994vietoris}
J.-C. Hausmann \emph{et~al.}, \emph{On the Vietoris-Rips complexes and a cohomology theory for metric spaces}.\hskip 1em plus 0.5em minus 0.4em\relax Universit{\'e} de Gen{\`e}ve-Section de math{\'e}matiques, 1994.

\end{thebibliography}


\end{document}